\documentclass[letterpaper]{article} 
\usepackage{arxiv-version}  
\usepackage{times}  
\usepackage{helvet}  
\usepackage{courier}  
\usepackage[hyphens]{url}  
\usepackage{graphicx} 
\usepackage{natbib}  
\usepackage{caption} 
\usepackage{algorithm}
\usepackage{algorithmic}
\usepackage{amsmath} 
\usepackage{comment}
\usepackage{newfloat}
\usepackage{listings}
\DeclareCaptionStyle{ruled}{labelfont=normalfont,labelsep=colon,strut=off} 
\floatstyle{ruled}
\newfloat{listing}{tb}{lst}{}
\floatname{listing}{Listing}
\title{Keeping Experts in the Loop: Expert-Guided Optimization for Clinical Data Classification using Large Language Models}
\author {
    Nader Karayanni\equalcontrib\textsuperscript{\rm 1},
    Aya Awwad\equalcontrib\textsuperscript{\rm 2}, \\
    Chieh-Lien Hsiao\textsuperscript{\rm 1},
    Surish P Shanmugam\textsuperscript{\rm 2}
}
\affiliations {
    \textsuperscript{\rm 1}Columbia University
    \textsuperscript{\rm 2}Harvard Medical School\\
    nader.karayanni@columbia.edu, Aya\_awwad@hms.harvard.edu, \\ ch3689@columbia.edu, surishpshanmugam@hms.harvard.edu
}

\usepackage{bibentry}

\newcommand{\fwname}{StructEase}
\newcommand{\samplingname}{SamplEase}

\begin{document}

\maketitle

\begin{abstract}
Since the emergence of Large Language Models (LLMs), the challenge of effectively leveraging their potential in healthcare has taken center stage. A critical barrier to using LLMs for extracting insights from unstructured clinical notes lies in the prompt engineering process. Despite its pivotal role in determining task performance, a clear framework for prompt optimization remains absent.
Current methods to address this gap take either a manual prompt refinement approach, where domain experts collaborate with prompt engineers to create an optimal prompt, which is time-intensive and difficult to scale, or through employing automatic prompt optimizing approaches, where the value of the input of domain experts is not fully realized. 
To address this, we propose \fwname, a novel framework that bridges the gap between automation and the input of human expertise in prompt engineering. A core innovation of the framework is \samplingname, an iterative sampling algorithm that identifies high-value cases where expert feedback drives significant performance improvements. This approach minimizes expert intervention, to effectively enhance classification outcomes. 

This targeted approach reduces labeling redundancy, mitigates human error, and enhances classification outcomes.
We evaluated the performance of \fwname\ using a dataset of de-identified clinical narratives from the US National Electronic Injury Surveillance System (NEISS), demonstrating significant gains in classification performance compared to current methods. Our findings underscore the value of expert integration in LLM workflows, achieving notable improvements in F1 score while maintaining minimal expert effort. By combining transparency, flexibility, and scalability, \fwname\ sets the foundation for a framework to integrate expert input into LLM workflows in healthcare and beyond. 

\end{abstract}

%

\section{Introduction}

With the advent of large language models (LLMs), the landscape of healthcare delivery is evolving as these technologies find novel applications across the field~\cite{epic2024gpt4,landi2024nuance}. Prompt engineering has emerged as an important element to harness the full potential of LLMs, where minor changes in the natural language inputs can impact the output's quality and relevance \cite{mesko2023prompt}.
Among the various applications, the classification of unstructured clinical narrative notes from electronic medical records (EMRs) is particularly relevant for researchers, healthcare administrators, and policymakers \cite{wang2018clinical}. Medical records are rich with clinical data that are not always fully captured by structured fields like International Classification of Diseases (ICD) codes, making the ability to navigate and classify unstructured free text essential.

The challenge lies in extracting this information effectively. Previous studies have shown that generating task-specific prompts typically requires the combined efforts of prompt engineers, who have a deep understanding of LLMs and best prompting practices, and domain experts, who are well-versed in the task-specific intricacies \cite{zhang2024evaluating, huang2024critical, guevara2024large, burford2024use, jethani2023evaluating}. This ensures creating an ideal prompt where elements such as task description, domain-specific details, and benchmarks are incorporated to drive the prompt quality and performance. However, refining prompts often involves an iterative, manual exploration process that is tedious, costly and requires technical expertise \cite{pryzant2023automatic}. Moreover, the absence of a systematic framework to guide this process increases inefficiencies and limits the wider adoption of these methods \cite{ding2021openprompt}.

To address these challenges, new methods are being developed to improve and automate prompt engineering. Beyond single-prompt techniques such as zero-shot and few-shot learning, meta-prompting approaches are gaining traction \cite{prompt_engineering_medium, metaPromptingGuide}. In essence, those approaches generate, refine, and optimize LLM prompts for a given task. Techniques such as evolutionary algorithm (e.g., Monte Carlo search, Gibbs sampling) and textual gradient-based optimization are some of the underlying methods to achieve this automation \cite{zhang2024evaluating,yuksekgonul2024textgrad}. While effective, these approaches limit user involvement in prompt optimization.

Of interest is the work introduced by PromptAgent \cite{wang2023promptagent}, which addresses the challenge of generating expert-level prompts. PromptAgent employs Monte Carlo Tree Search (MCTS) to navigate the prompt space strategically, mimicking human experts to achieve nuanced and high-quality prompts. While PromptAgent effectively highlights the limitations of prior methods and advances automated prompt engineering, it relies on an exhaustive exploration of the expert prompt space. It introduces practical constraints, where the process can take hours to iterate through the search space—inherently lacking domain-specific knowledge and feedback during runtime. 
Similarly, DSPy \cite{khattab2022demonstrate, khattab2024dspy}, provides a programmatic framework for managing LLM interactions and refining prompts, abstracting the challenges of prompt engineering from programmers, and automatically generating the task-specific prompts with the best practices of prompt engineering. 

The strive to automate the prompt engineering process is understandable, given the tedious nature of the task. However, the cost of excluding the expert from the loop can be significant. Recent work in material science highlights the critical role of human expertise in understanding AI-generated results, emphasizing that while AI systems excel at computation and automation, human expertise provides the vision and contextual understanding necessary to make the results meaningful \cite{toner2024artificial}. 

Building on this understanding, we introduce \fwname, an innovative framework that \textbf{bridges the gap between full automation and expert involvement in prompt engineering}. Designed to classify large sets of unstructured clinical notes using LLMs, \fwname\ keeps domain experts at the center of the workflow, ensuring iterative refinement of classification prompts through expert-guided corrections. Unlike fully automated approaches, \fwname\ strategically integrates expert feedback into the prompt generation process, while addressing key challenges such as class imbalances and redundant sampling. Our novel approach ensures that expert feedback effectively distills domain knowledge into the system while requiring minimal effort from the expert. Our research shows that leveraging an expert in the loop outperforms the fully automated approaches and provides a novel approach to do so effectively.

\section{Methods}
\subsection{Proposed Framework}
\begin{figure}[t]
\centering
\includegraphics[width=1\columnwidth]{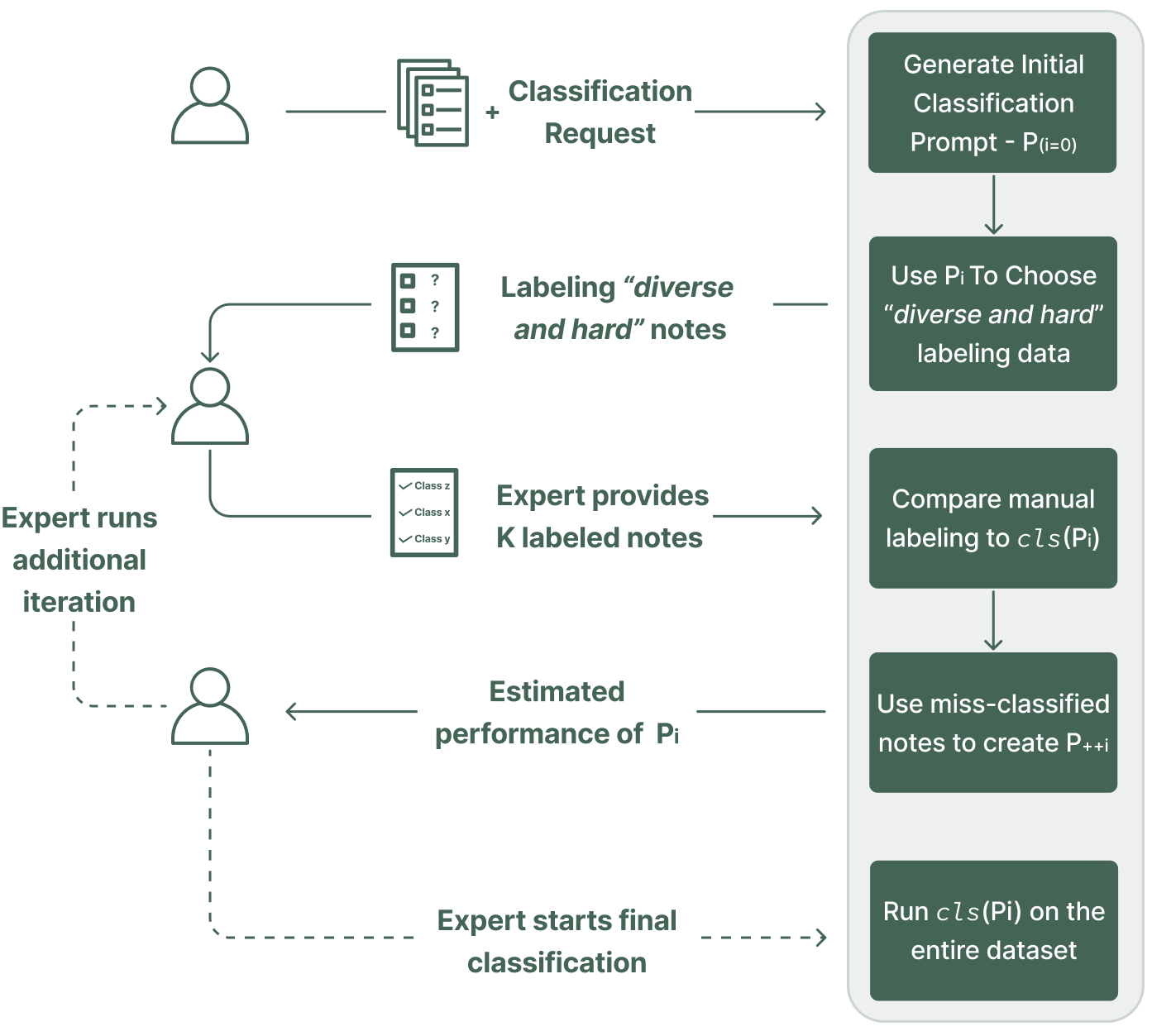}
\caption{The proposed framework of \fwname}
\label{Figure 1}
\end{figure}

\vspace{0.1cm}
We propose \fwname, a novel framework that integrates domain experts into the AI workflow to enable the classification of large sets of unstructured clinical notes using LLMs. The framework is designed to efficiently distill domain expertise into classification prompts, which are iteratively refined to improve the performance and relevance of LLM outputs. An overview of the framework is illustrated in Figure~\ref{Figure 1}.

\subsubsection{Classification Prompt Generation}
At the core of \fwname\ is a classification prompt's generation and iterative refinement. This prompt is generated and refined to fulfill the clinical note classification request. The prompts are designed using Chain-of-Thought (CoT) reasoning to enhance the effectiveness of classification tasks. \fwname\ starts by creating an initial classification prompt $P_0$ and continuously improves it after each labeling iteration. The prompt is improved by refining the classification instructions and incorporating Few-Shot (FS) examples from expert-labeled data. We annotate the prompt generation methods as follows:
\[
P_0 = GenerateInitialPrompt(Classes, Request, Dataset)
\]
\[
P_i = UpdatePrompt(P_{i-1}, Few\_Shots)
\]
\textit{Few\_Shots} are the notes that the prompt \(P_{i-1}\) classified incorrectly compared to the expert labels. The above methods use an LLM call to generate the new prompt.\\

\subsubsection{Choosing Data for Expert Review} A critical challenge in designing \fwname\ is selecting data samples that maximize the value of expert-labeled data. This is particularly difficult in scenarios with class imbalances, where it is essential to ensure that the \textbf{reviewed data samples adequately represent all classes}. Another challenge arises from the possibility of a large proportion of relatively simple samples in the dataset, which may not contribute meaningfully to improving the model’s classification performance. Furthermore, it is crucial to \textbf{avoid redundant expert reviews across labeling iterations} to ensure that each iteration yields unique and valuable insights that enhance the classification capabilities of the LLM. To address these challenges, we propose \samplingname, A novel algorithm, detailed in Algorithm~\ref{iterative_labeling}. \\
\begin{algorithm}
\caption{\samplingname}
\label{iterative_labeling}
\textbf{Input:} Dataset $D$, Classification\_Classes $\textit{Classes}$, Classification\_Request $\textit{Request}$ \\
\textbf{Output:} Labeled dataset $\textit{labeled\_data}$\\
\begin{algorithmic}[1]
\STATE $\textit{labeled\_data} \gets \emptyset$, $i \gets 0$
\STATE $P_i \gets \textsc{GenerateInitialPrompt}(\textit{Classes}, \textit{Request}, \textit{D})$
\WHILE{the expert opts to continue labeling}
    \STATE $S \gets \textsc{RandomSample}(D \setminus \textit{labeled\_data}, 10\%)$
    \STATE $C \gets \textsc{Classify}(S, P_i)$
    \STATE $conf \gets \textsc{ComputeConfidence}(C)$
    \FOR{each class $c \in \textit{Classes}$}
        \STATE $S_c \gets \textsc{SelectLowestConfidence}(C, conf, c, 10)$
    \ENDFOR
    \STATE $\textit{samples\_to\_label} \gets \bigcup_{c \in \textit{Classes}} S_c$
    \STATE Query the expert to label $\textit{samples\_to\_label}$
    \STATE $\textit{labeled\_data} \gets \textit{labeled\_data} \cup \textit{samples\_to\_label}$
    \STATE $\textit{few\_shots} \gets \textsc{Mismatched}(P_i, \textit{samples\_to\_label})$
    \STATE $i \gets i + 1$
    \STATE $P_{i} \gets \textsc{UpdatePrompt}(P_i, \textit{few\_shots})$
\ENDWHILE
\STATE Return classification of $D$ using $P_i$
\end{algorithmic}
\end{algorithm}

\textsc{ComputeConfidence} calculates the confidence of the LLM for classifying each clinical note separately as follows: 
\begin{equation}
    \text{Confidence} = \exp\left(\frac{1}{n} \sum_{i=1}^{n} \log P_i \right)
\end{equation}
Where:
\begin{itemize}
    \item \( n \) is the number of tokens in the completion.
    \item \( \log P_i \) is the log probability of the \(i\)-th token.
    \item \( \exp(x) \) is the exponential function, \( e^x \).
\end{itemize}

\vspace{10pt} 
Through this iterative process, \fwname\ continuously incorporates expert insights to improve the classification prompt. At any stage, the expert can deploy the most recent prompt, applied across the entire dataset to classify the clinical notes. By abstracting the technical complexities, \fwname\ is designed to enhance classification accuracy and to ensure advanced AI capabilities are accessible to domain specialists with minimal technical training.
\subsubsection{Framework Implementation}
We developed \fwname\ in Python and it is available as an open-source project on GitHub\footnote{\url{https://github.com/karayanni/StructurEase/}}. The implementation includes a web-based application that provides a user-friendly interface, enabling users to upload datasets, define classification requests, and engage in the iterative labeling process.
To meet real-world needs, the implementation incorporates two key steps. First, the classification process using $P_i$ is parallelized across clinical notes to minimize latency during labeling iterations. This parallelization significantly reduces the time required for both iterative refinements and the final dataset classification. Furthermore, the end-to-end framework is containerized using Docker \cite{merkel2014docker}, to encourage adoption and future research and collaboration.

\subsection{Data and Task Definition} 
\vspace{0.1cm}
The data for this study was sourced from the 2023 US Consumer Product Safety Commission's National Electronic Injury Surveillance System (NEISS). This system operates on a nationally representative, stratified probability sample from 96 hospitals across the US and its territories, each equipped with at least six beds and an emergency department (ER). NEISS provides publicly accessible, de-identified data that do not require institutional review board approval or informed consent for use, in compliance with the HIPAA Privacy Rule. We filtered for micromobility-related visits based on NEISS product codes. The final dataset included 17,888 different clinical narratives \cite{cpsc2024neiss, burford2024use}.  \\

In Our study, we use the "gpt-4o-mini-2024-07-18" model accessible via the OpenAI API \cite{gpt4o-mini}. To ensure reproducibility and consistency, we configured the language model with a temperature setting of 0 and a top-p value of 1. This setup ensures that the model consistently selects the most probable token during text generation.
As a use case, the specific task was to extract helmet usage status from unstructured clinical narratives of ER visits. The goal was to classify each note into one of three classes: helmet, no helmet, and cannot determine helmet status. To evaluate our framework, 2,000 clinical narratives were randomly selected and independently reviewed by two expert physicians. Any disagreements between the reviewers were resolved through discussion until a 100\% inter-rater agreement was reached. It is important to note that the manually labeled dataset created for this study is proprietary and has not been published online, addressing any potential concerns regarding data leakage to the LLM before this study.

\subsection{Evaluation Experiments}
\vspace{0.1cm}
We conducted a series of experiments to evaluate our framework under various conditions:

\begin{figure*}[t]
\centering
\includegraphics[width=1\textwidth]{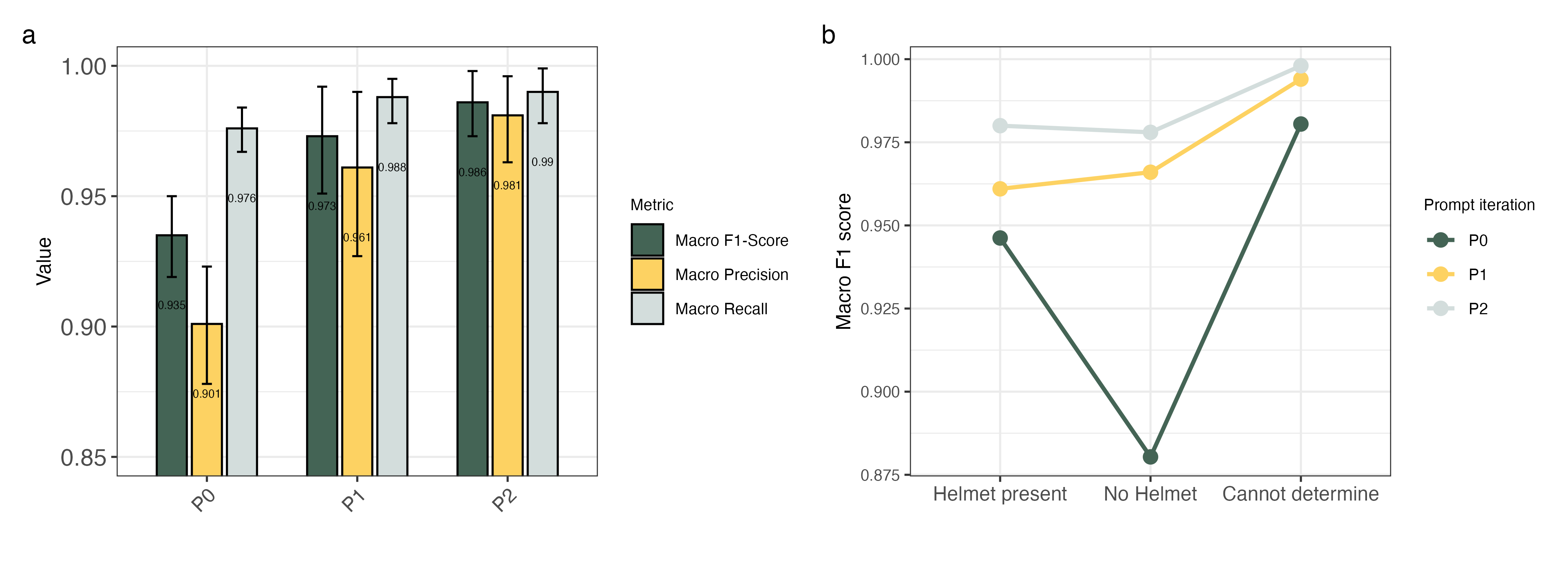} 
\caption{Aggregated performance metrics across iterative prompt refinements (\(P_0\), \(P_1\), \(P_2\)). (\textbf{a}) Overall macro-level metrics (Macro F1-Score, Macro Precision, and Macro Recall), showing consistent improvements with each iteration of the prompt. Error bars indicate 95\% confidence intervals. (\textbf{b}) Per-class performance analysis of Macro F1-Score for the "Helmet present," "No Helmet," and "Not mentioned" categories. Iterative refinements yielded performance improvements, particularly for the "No Helmet" class, which had the lowest initial performance at \(P_0\). }
\label{fig2}
\end{figure*}

\begin{enumerate}
    \item \textbf{Iterative Prompt Refinement:} We evaluated the performance of \fwname\ by running the framework six times independently, generating the prompts \(P_0\), \(P_1\), and \(P_2\) in each run. Starting with the baseline prompt \(P_0\), the framework iteratively refined the prompts through expert feedback to produce \(P_1\) and \(P_2\). The experiment aimed to assess how iterative prompt refinement improves classification performance across runs.
    
    \item \textbf{Comparing \samplingname\ vs. Random Sampling:} To evaluate the sampling approach used in \fwname, we compared \samplingname\ with a baseline random sampling method. Six independent runs were conducted for each sampling strategy using a single prompt iteration (\(P_1\)). This experiment aimed to explicitly assess the novel proposed sampling algorithm in improving the performance of the framework compared to random sampling.

    \item \textbf{Comparison with Baselines:} We compared the performance of \fwname\ against four baseline methods designed for similar classification tasks: (1) Human prompt: human-provided instructions for direct classification task. (2) DSPy Chain of Thought (CoT) classifier: A DSPy classifier that incorporates step-by-step reasoning to improve interpretability and accuracy. (3) DSPy optimized: A version of the DSPy CoT classifier that requires labeled data.~\cite{khattab2022demonstrate, khattab2024dspy}. This method utilizes the provided labeled data to create a CoT classification prompt and then fine-tunes the LLM to minimize the loss for the classification task. Although this approach involves fine-tuning---unlike the other methods---we included it to assess how effective our approach can be when focusing solely on prompt optimization. We ran this version with 30 randomly labeled samples to benchmark its performance.
    
    
    
    
    
     
    \item \textbf{Bias evaluation:} To examine potential biases in model performance, we evaluated the framework across key demographic groups. Specifically, we compared the model's performance across genders (males vs. females) and across self-reported racial categories: White, Black/African American, Asian, Other, or Not Specified.
\end{enumerate}

Unless otherwise mentioned, all our experiments are conducted with the \samplingname\ default parameter, specified in Algorithm~\ref{iterative_labeling}.

\subsection{Performance Metrics and Statistical Analysis}
\vspace{0.1cm}
\subsubsection{Overall and Per-Class Evaluation}
The framework was evaluated using 2,000 expert-labeled notes (ground truth), excluding the training data used to refine the prompt. Performance was assessed using macro-level classification metrics (macro-precision, macro-recall, and macro-F1 score) and per-class metrics (precision, recall, and F1 score). These metrics were defined as follows:

\begin{itemize}
\item Precision = TP / (TP + FP)
\item Recall = TP / (TP + FN)
\item F1 = (2 $\cdot$ Precision $\cdot$ Recall) / (Precision + Recall)
\\ {\footnotesize \textit{(Abbreviations: TP = true positives, FP = false positives, FN = false negatives)}}
\end{itemize}

The framework was evaluated across six independent runs to account for variability in the model outputs. Macro-level metrics were computed by averaging the class-specific metrics, while per-class metrics were calculated separately for each class.

To quantify uncertainty, 95\% confidence intervals (CI) were derived using bootstrap resampling (n=1000). For macro-level metrics, resampling was performed within each run, and the resulting bootstrap distributions were aggregated across the six runs to compute final CIs.

For per-class metrics, stratified bootstrap resampling was applied to preserve class proportions within each resampled dataset. Metrics for each class were recalculated for every bootstrap iteration, and results were aggregated across runs. Final point estimates and 95\% confidence intervals were reported for both overall and per-class metrics. \\

\subsubsection{Statistical Analysis}
To compare Macro F1-Score between prompt iterations (P$_0$ vs. P$_2$), we calculated the mean difference and its 95\% confidence interval using the bootstrap distributions. Statistical significance was determined by whether the confidence interval excluded zero. A permutation test further validated the difference by generating a null distribution of shuffled samples. 
For the sampling experiments, the medians of the performance metrics across the runs were calculated. Differences in these medians between the sampling methods were compared using the Mann–Whitney U test, a non-parametric test for comparing medians between two independent groups. A two-sided significance threshold of $P \leq 0.05$ was used. All statistical analyses and visualization were conducted using R software (version 4.4.1).

\section{Results}
\subsection{Iterative Prompt Refinement}
\vspace{0.1cm}
Figure~\ref{fig2} presents the aggregated results of the iterative prompt refinement process. The iterative refinement of prompts (\(P_0 \to P_1 \to P_2\)) resulted in consistent improvements in overall (Figure~\ref{fig2}a) and per-class performance metrics (Figure~\ref{fig2}b). F1 score increased from 0.935 (95\% CI: 0.919–0.951) with the baseline prompt (\(P_0\)) to 0.973 (95\% CI: 0.951–0.992) after the first refinement (\(P_1\)) and reached 0.986 (95\% CI: 0.972–0.997) after the second refinement (\(P_2\)). The increase in F1 score and precision between (\(P_0 \to P_2 \)) was statistically significant with \(P < 0.001\). Similar trends were observed for precision (0.901 (95\% CI: 0.879-0.923) at \(P_0\) to 0.981 (95\% CI:0.963-0.996) at \(P_2\)) and recall (0.976 (95\% CI: 0.967-0.984) at \(P_0\) to 0.990 (95\% CI: 0.979-0.999) at \(P_2\)).

Per-class analysis of F1 scores (Figure~\ref{fig2}b) shows significant improvements across all categories, particularly for the "No Helmet" class, where initial performance (\(P_0\)) was the lowest at 0.88 (95\% CI: 0.85, 0.91), but increased markedly to 0.98 (95\% CI: 0.95-1) at \(P_2\)), this $\Delta 10\%$ improvement was statistically significant \((P < 0.001\)).

\subsection{Comparing \samplingname\ vs. Random Sampling}
\vspace{0.1cm}

\begin{figure}[t]
\centering
\includegraphics[width=1\columnwidth]{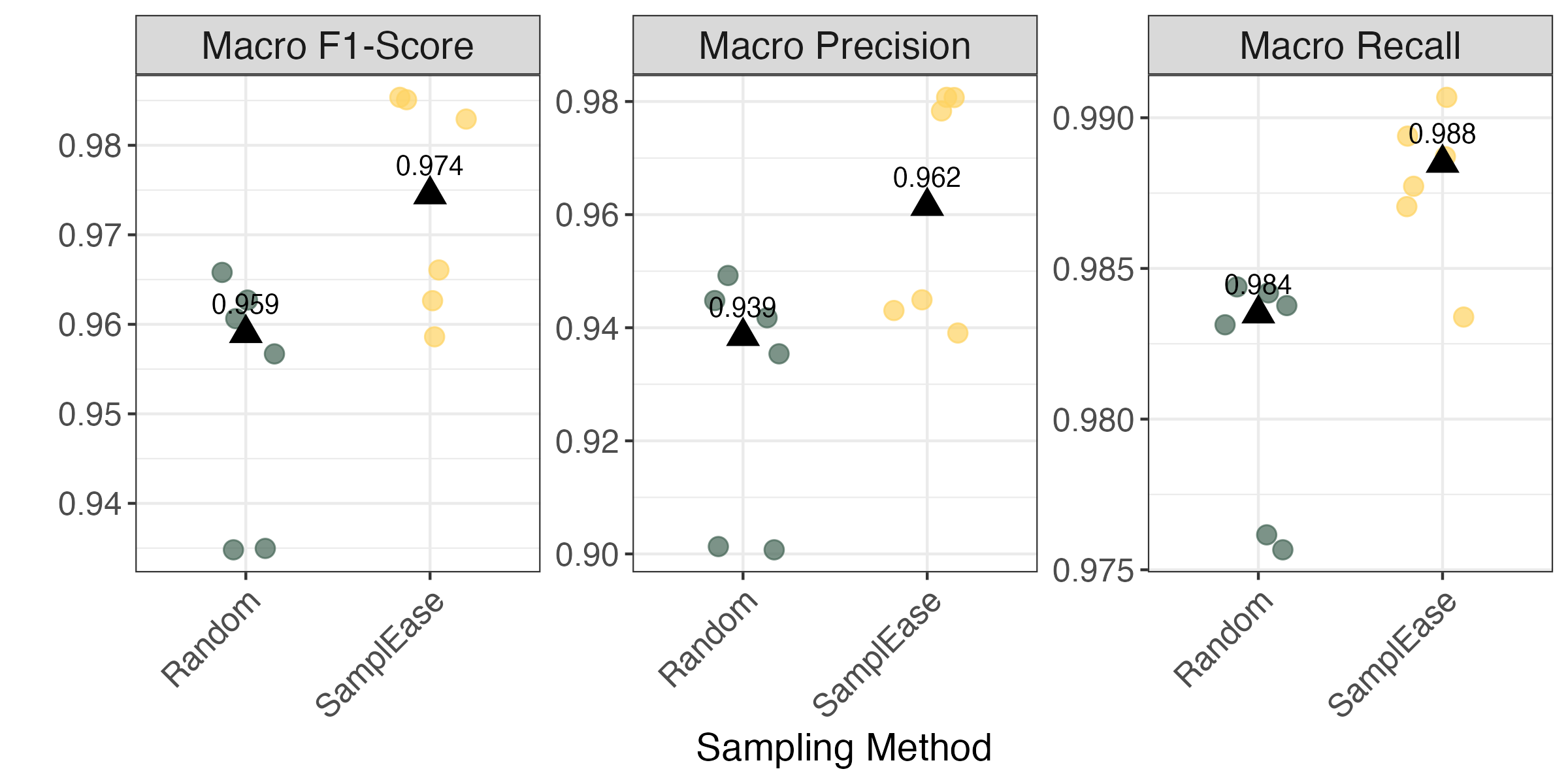} 
\caption{Comparison of performance metrics between smart sampling and random sampling strategies. Each point represents a performance metric (Macro F1-Score, Macro Precision, Macro Recall) from six independent runs for each sampling method. Triangles indicate the median performance for each metric within the sampling method.}
\label{Figure 3}
\end{figure}

Figure~\ref{Figure 3} presents the performance comparison between smart and random sampling strategies across six independent runs. Smart sampling demonstrated consistent improvements in performance metrics compared to random sampling. For the F1 score, smart sampling achieved a higher median value (0.974) compared to random sampling (0.959), with a statistically significant difference (\(P = 0.044\)). Similarly, recall showed a significant improvement, with smart sampling yielding a median value of 0.988 compared to 0.984 for random sampling (\(P = 0.023\)). The improvement in precision (0.962 for smart sampling vs. 0.939 for random sampling) did not reach statistical significance (\(P = 0.077\)).

\subsection{Comparison with Baselines}
\vspace{0.1cm}

\begin{table*}[ht]
\centering
\renewcommand{\arraystretch}{1.2} 
\setlength{\tabcolsep}{6pt} 
\begin{tabular}{lcccc}
\hline
 & \textbf{Macro F1-Score} & \textbf{Macro Precision} & \textbf{Macro Recall} & \textbf{Accuracy} \\ \hline
\textbf{Human prompt}  & 0.735 & 0.753 & 0.865 & 0.730 \\
\textbf{DSPy CoT classifier}         & 0.962 & 0.953 & 0.972 & 0.981 \\
\textbf{DSPy Optimized}   & 0.980 & \textbf{0.982} & 0.978 & 0.991 \\
\textbf{P$_1$}            & 0.973 & 0.961 & 0.988 & 0.988 \\
\textbf{P$_2$}            & \textbf{0.986} & 0.981 & \textbf{0.990} & \textbf{0.995} \\ \hline
\end{tabular}
\caption{Performance comparison of different approaches across Macro F1-Score, Macro Precision, Macro Recall, and Accuracy. Bold values indicate the highest performance in each metric. \textit{Abbreviations:} CoT, chain of thought; P$_1$, first prompt iteration of \fwname; P$_2$, second prompt iteration of \fwname.}
\label{table:performance_comparison}
\end{table*}

The results in Table~\ref{table:performance_comparison} demonstrate that \fwname\ mostly outperformed other baseline methods across the evaluated metrics. The second prompt iteration P$_2$ achieved the highest performance, with a F1-Score of 0.986, precision of 0.981, recall of 0.990, and accuracy of 0.995.

Among the baselines, DSPy Optimized showed strong performance, with F1-Score of 0.980 and an accuracy of 0.991. However, \fwname\ surpassed these results, particularly in recall and F1-Score, highlighting the advantage of iterative prompt refinement. By contrast, the human prompt achieved the lowest overall performance, with a F1-Score of 0.735 and accuracy of 0.730.

\subsection{Bias Evaluation}
\vspace{0.1cm}
The bias evaluation experiment demonstrated that the framework performed consistently across different demographic groups, with minimal expected variation in classification metrics between genders and among self-reported racial categories. When comparing performance across genders (male vs. female), the results indicated no substantial disparities in macro-level metrics or per-class evaluations. Similarly, when examining performance across racial categories (White, Black/African American, Asian, Other, Not Specified), the framework maintained equitable classification outcomes across all groups.

\section{Conclusion and Future Work}
Effective prompt engineering for leveraging LLMs in healthcare-related tasks is pivotal, yet current research falls short in bridging automation with domain expertise. This gap arises from the absence of established frameworks and the challenge of integrating human expertise strategically rather than mimicking it entirely.

To address this gap, we present \fwname, a novel framework that empowers domain experts to perform a classification of clinical notes without needing prior expertise in LLMs. By leveraging an iterative process of minimal yet impactful expert input, \fwname\ demonstrates how strategic integration of domain expertise can significantly enhance system performance.

\fwname\ achieves substantial improvements in using LLMs for both overall and per-class classification performance. A critical challenge our framework addresses is identifying data samples where expert input yields the greatest performance gains. To this end, we introduce \samplingname, a novel algorithm for iterative sampling that consistently outperforms random sampling. Our results demonstrate significant gains in F1 score and recall respectively, underscoring \samplingname’s efficiency in leveraging expert input.

By focusing the expert attention to a select sample of the data, \fwname\ reduces labeling errors stemming from fatigue or ambiguity in simpler, repetitive tasks. Even with well-defined rules and training, labeling variability is common. Our framework mitigates these challenges by selecting high-value data for expert review, ensuring consistent improvement with each iteration \cite{wong2022ground}.

In our work, we optimized the classification performance solely by refining the prompt $P_i$. Other methods to further improve classification quality exist, such as dynamically fine-tuning the LLM---as demonstrated by the DSPy Optimized method in our baseline comparisons. Exploring such additional techniques to enhance \fwname\ remains an exciting avenue for future research. Our evaluation shows that integrating expert input through \samplingname\ outperforms the more sophisticated LLM optimization methods that utilizes random labeled samples for their refinement, highlighting the value of our approach in \samplingname. The performance advantage is expected to improve even further with the integration of additional refinement techniques.

A key challenge in designing \fwname\ is minimizing expert effort while ensuring continuous improvement. To achieve this, \samplingname\ limits the labeling workload per iteration to a manageable $10 \times |Classes|$. The framework also offers users flexibility by allowing them to determine the number of iterations required to refine $P_i$, accommodating diverse accuracy requirements.
Transparency is another core strength of our framework. At every stage, users see how their input shapes the prompts, fostering trust and bridging the gap between domain expertise and LLM workflows. Over time, this transparency could empower users to better understand LLMs and critically evaluate their outputs. This represents an interesting direction for future research, particularly in evaluating how such collaborative frameworks can empower domain experts to critically assess LLM outputs, with significant implications for the design of human-AI collaborations~\cite{jabbour2023measuring}.


Even though the classification task we presented did not explicitly explore non written information, we examined the performance across gender and different racial groups. There was no evidence of disparities in the overall or per-class performance within those groups. This consistency likely reflects the nature of the classification task. However, as LLMs are applied to different tasks, examining the fairness of the model is warranted~\cite{duan2024large}.

\fwname\ was initially designed and evaluated for the specific task of dynamically classifying unstructured clinical notes. However, its methodology—integrating expert feedback iteratively—can be generalized across a variety of tasks and LLM workflows. By maintaining a consistent focus on efficiently incorporating expert input, \fwname\ can be adapted to other downstream tasks requiring similar expert-in-the-loop interactions, further broadening its applicability.

When deployed in real-world scenarios involving identifiable clinical data, privacy, and regulatory compliance become critical considerations. To ensure HIPAA compliance, it is essential to utilize LLMs that meet stringent security and privacy standards. Additionally, the framework must be designed to restrict data access and iterative processes exclusively to authorized users, safeguarding sensitive information throughout the workflow. Addressing these challenges, including adapting \fwname\ to operate within secure, locally hosted, or private LLM environments, will be a key focus of future research and development.

\bibliography{Arxiv-Version}
\end{document}